\documentclass[conference]{IEEEtran}
\usepackage{multirow}

\usepackage{dblfloatfix}

\ifCLASSINFOpdf
  \usepackage[pdftex]{graphicx}
  \usepackage[export]{adjustbox}
  \usepackage{tabularx,booktabs}
  \newcolumntype{C}{>{\centering\arraybackslash}X} 
  \usepackage{multirow}
\else
\fi

\usepackage{amsmath}
\usepackage{amssymb}
\usepackage{tikz, pgfplots}
\usepackage{pgfplotstable}
\usetikzlibrary{positioning, matrix, tikzmark}
\pgfplotsset{width=9.5cm}
\usepackage{capt-of}
\usepackage{float}
\usepackage[hidelinks]{hyperref}
\usetikzlibrary {arrows.meta}
\usetikzlibrary{external}
\begin{document}

\title{The Sparse Tsetlin Machine: \\ Sparse Representation with Active Literals}

\author{
\IEEEauthorblockN{Sebastian Østby}
\IEEEauthorblockA{Forsta Labs\\
Center for AI Research\\
University of Agder\\
Grimstad Norway\\
sebastiaos@uia.no\\
0000-0002-1159-2430}
\and
\IEEEauthorblockN{Tobias M. Brambo}
\IEEEauthorblockA{Forsta Labs\\
Center for AI Research\\
University of Agder\\
Grimstad Norway\\
tobiasmb@uia.no\\
0000-0002-7838-0467}
\and
\IEEEauthorblockN{Sondre Glimsdal}
\IEEEauthorblockA{Forsta Labs\\
Center for AI Research\\
University of Agder\\
Grimstad Norway\\
sondre.glimsdal@forsta.com\\
0000-0001-9735-4064}
}

\maketitle

\begin{abstract}

This paper introduces the Sparse Tsetlin Machine (STM), a novel Tsetlin Machine (TM) that processes sparse data efficiently. Traditionally, the TM does not consider data characteristics such as sparsity, commonly seen in NLP applications and other bag-of-word-based representations. Consequently, a TM must initialize, store, and process a significant number of zero values, resulting in excessive memory usage and computational time. Previous attempts at creating a sparse TM have predominantly been unsuccessful, primarily due to their inability to identify which literals are sufficient for TM training. By introducing Active Literals (AL), the STM can focus exclusively on literals that actively contribute to the current data representation, significantly decreasing memory footprint and computational time while demonstrating competitive classification performance. 

\end{abstract}

\bigskip

\IEEEpeerreviewmaketitle

\section{Introduction}
The Tsetlin Machine (TM) employs groups of Tsetlin Automata (TA) in conjunctive clauses to formulate logical rules, capturing sub-patterns of data to perform complex pattern recognition \cite{granmo2021tsetlin}. Traditionally, Artificial Neural Network facilitates online learning of non-linear patterns by minimizing output error, frequently resulting in training overfitting. In contrast, a TM leverages principles of frequent pattern mining and resource allocation to extract prevalent patterns. This, in turn, reduces overfitting while also making clauses transparent by having interpretable self-contained patterns that can be expressed as logical rules, such as:

\begin{quote}
    \textbf{If} Condition A \textbf{and not} Condition B \textbf{then} Result C.    
\end{quote}

The TM yields competitive results compared to other classification methods when performing benchmarks, and the hardware-near approach reduces energy footprint and minimizes memory impact during training and inference \cite{bhattarai2023contracting}. TMs encompass a diverse range of machine learning paradigms, including classification\cite{granmo2021tsetlin}, convolution\cite{granmo2019convolutional}, regression\cite{darshana2020regression}, and reinforcement learning \cite{rahimi2023off}. Recent research has harnessed TMs for various applications, such as sentiment analysis  \cite{yadav2021human}, fake news detection  \cite{bhattarai2021explainable}, and novelty detection \cite{bhattarai2021wordlevel}\cite{bhattarai2020measuring}. TMs have also been applicable in the healthcare sector \cite{berge2019using}.

Traditionally, a TMs efficiency decreases as a problem's complexity increases. In addition, a TM does not consider inherent data characteristics, such as sparsity, commonly seen in Natural Language Processing (NLP) and other bag-of-word-based representations. In NLP applications, binary encodings that frequently result in sparse representations are required for TM training. Consequently, memory must be initialized, stored, and processed for an extensive feature set of mostly zero values, accumulating in excessive and inefficient computational time. Clearly, a new TM that better handles sparsely populated data, with compact data compression and more efficient processing, should result in an accurate sparse representation for the TM.

This paper presents the Sparse Tsetlin Machine (STM), a novel TM that introduces Active Literals (AL) to leverage sparsity. The AL acts as a gatekeeper at a class level, introducing literals and sub-patterns from the data into clauses via frequent pattern mining. This enables effective sparse data processing. The STM will only distribute resources to discriminative literals, ensuring that only literals that actively influence predictions are considered. Novel to the STM is the ability to initialize the TM memory with empty clauses that contain no record of TAs at the beginning of training. Then, through feedback, the AL introduces literals and sub-patters contained in the data into memory. Further, a lower-bound threshold creates a minimum TA state allowed in clauses. This establishes an allowed state spectrum where TAs can fluctuate. If a state is reduced below the allowed lower bound, the TA is removed from the clause immediately, making the memory smaller and patterns contained in clauses robust and concise. This coalesces in a new efficient sparse memory space for the STM that efficiently performs memory feedback, distributing resources solely to literals that actively contribute to the current data representation, dramatically increasing the processing capabilities of sparse data. The STM is extensively evaluated on a vast amount of well-known datasets. When processed for TM training, these datasets become vastly sparse and are ideal for confirming the STM's ability to process sparse data while achieving competitive performance efficiently. Furthermore, the compact memory usage of the STM permits traditional TM training on large-scale datasets, such as big text corpora, a previously unattainable capability.

\section{Tsetlin Machines}
The TM uses conjunctive clauses, composed of two-action TAs, to solve complex pattern recognition. Each TA has $2N$ states, where the first N states represent the \textit{Exclude} action, and the rest $N+1 \text{ to } 2N$ states represent the \textit{Include} action. The states for a TA are updated through iterative feedback \cite{granmo2021tsetlin}. The general structure of the TM can be thought of as a triplet of the input, output, and memory space, shown in Eq.~\ref{eq:reg_tm_structure}, following the notation from \cite{glimsdal2021coalesced}. 

\begin{equation}
    \label{eq:reg_tm_structure}
    \{ \mathcal{X}, \mathcal{Y}, \mathcal{C} \}
\end{equation}

The \textbf{input space} ($\mathcal{X}$) of the TM consists of input vectors $x$ for $o$ propositional inputs: $x = \left[x_{1},\ldots,x_{o}\right] \in \mathcal{X}, \mathcal{X} = \{0,1\}^{o}$. The TM requires this boolean representation of the input data.

The \textbf{output space} ($\mathcal{Y}$) in the TM consists of two possible output values: $y \in \mathcal{Y}, \mathcal{Y} = \{0,1\}$. This entails that a singular TM can only decide between two classes at most, meaning that for problems with multiple classes multiple TMs are required to work together, where each TM is assigned its own output to represent.

The \textbf{memory space} $\mathcal{C} = \{1,2,\ldots,2N\}^{n \times 2o}$ consists of the clause matrices $C$. $N$ defines the number of include and exclude states in the TM, and $n$ is a user-defined parameter that sets the number of clause rows in the matrix. The columns represent the $o$ inputs in $x$, along with their negations, referred to as literals: $L = \left[x_{1},\ldots,x_{o},\lnot x_{1},\ldots,\lnot x_{o}\right] = \left[l_{1},\ldots,l_{2o}\right]$. Each clause-row in the TM is also assigned a polarity $p \in \{+, -\}$, used in the majority voting phase. This polarity defines which output the clause should vote for, positive polarity votes for $y=1$ and negative polarity votes for $y=0$. There are an even number of positive and negative polarity clauses in the TM. 

From this, all clause-rows $C_{i}^{p} \in C$, being conjunctive, can be defined like in Eq.~\ref{eq:conjunctive_clause_regTM}.

\begin{equation}\label{eq:conjunctive_clause_regTM}
    C_{i}^{p}(x) = \bigwedge_{l_{k} \in L} l_{k}
\end{equation}

A given clause will output $1$ if, and only if, the input $x$ and the clause match completely, otherwise the output is $0$. After all the clauses have been evaluated on the input, the clause outputs are combined through summation and thresholded using the unit step function $u(v) = 1 \textbf{ if } v \geq 0 \textbf{ else } 0$ as shown in Eq.~\ref{eq:yhat_regTM}, where the first $\frac{N}{2}$ clauses represent positive polarity, and the rest negative.

\begin{equation}\label{eq:yhat_regTM}
    \hat{y} = u\left( \sum_{i=1}^{N/2} C_{i}^{+}(x) - \sum_{j=1}^{N/2} C_{j}^{-}(x) \right)
\end{equation}

This then gives a singular value for $\hat{y} \in \{0, 1\}$, based on the collective votes from the TM clauses.

\subsection{Coalesced Tsetlin Machine} \label{section:coalesced_description}

The Coalesced TM (CoTM) builds upon the regular TM by introducing the weight space $\mathcal{W}$. The CoTM structure is defined as a quadruple in Eq.~\ref{eq:coalesced_structure}. Though the input space ($\mathcal{X}$) remains the same between the two TMs, the introduction of the weight space provides significant differences in the way we define the other structural elements in the TM \cite{glimsdal2021coalesced}.

\begin{equation}
    \label{eq:coalesced_structure}
    \{ \mathcal{X} , \mathcal{Y} , \mathcal{C} , \mathcal{W} \}
\end{equation}

The \textbf{output space} ($\mathcal{Y}$) of the CoTM contains output $y$ for $m$ propositional outputs: $y \in \mathcal{Y}, \mathcal{Y} = \{0,\ldots,m-1\}$. In contrast to the regular TM, the necessity for multiple TMs to handle multiple outputs no longer exists, as the CoTM can handle multiple outputs using the same clauses through the weight matrix $W$.

The \textbf{memory space} $\mathcal{C}$ of the CoTM is very similar to that of the regular TM. However, the need for clause polarity no longer exists here, as the clauses can vote for multiple classes through the weight matrix $W$. 

The \textbf{weight space} $\mathcal{W} = \{\ldots, -2, -1, 0, 1, 2, \ldots\}^{m \times n}$ consists of the weight matrices $W$. Each clause matrix $C$ has a given weight matrix $W$, where each clause $C_{i} \in C$ is assigned weights for each output $y_{m} \in \mathcal{Y}$. For a positive weight, the clause is assigned the output value $1$ for the given $y_{m}$, and reversely $0$ for a negative weight. This allows for each clause to be assigned multiple outputs.

Like with the regular TM the conjunctive clauses can be defined from Eq.~\ref{eq:conjunctive_clause_regTM}, but without the polarity parameter. This entails that the votes now are calculated using the weight matrix $W$ and the clause outputs, given by Eq.~\ref{eq:majority_voting_cotm}. This then gives the total votes from all clauses for all $m$ classes.

\begin{equation}\label{eq:majority_voting_cotm}
    v = Wc
\end{equation}

Clauses receive feedback stochastically based on Eq.~\ref{eq:update_prob_coalesced}. Here, the update probabilities are calculated for each class $y_m \in y$ by comparing the vote sum $v_{i}$ and the summation margin $q = yt - \Bar{y}t$. The vote sum is clipped to be between $(-T, T)$, where $T$ is a user-defined threshold parameter defining how many clauses should vote for each output. This gives $m$ update probabilities $d$, one for each output class.

\begin{equation}\label{eq:update_prob_coalesced}
    d = 
    \begin{bmatrix}
    \multicolumn{1}{c}
    {|\frac{q_1 - clip(v_1, -T, T)}{2\cdot T}|}                    \\
            \vdots          \\
    |\frac{q_m - clip(v_m, -T, T)}{2\cdot T}|                    
    \end{bmatrix}
    = \begin{bmatrix}
            \multicolumn{1}{c}
            {d_1} \\
            \vdots \\
            d_m      
    \end{bmatrix}
\end{equation}

Feedback in the CoTM is given stochastically for each input pair $\left(x_{i},y_{i}\right)$ in the data with update probability $d_{i}$ and for a random class $\lnot y_{i}$ with probability $d_{k}$. If a clause should receive feedback for the corresponding class, it receives either Type Ia, Type Ib or Type II feedback.

\textbf{Type Ia Feedback} is given only if the input $x$ matches the clause, and that clause is assigned the output value $y_{i}$. Type Ia feedback then affects the True literals, reinforcing the \textit{Include} actions to make the clause more closely represent the input $x_{i}$.

\textbf{Type Ib Feedback} is given when the clause output value is assigned to $\lnot y_{i}$, and the clause \textit{doesn't} match the input $x_{i}$. This then coarsens the clause by reinforcing \textit{Exclude} actions, making the clause forget literals.

\textbf{Type II Feedback} is, like with Type Ib, given when the clause output value is assigned to $\lnot y_{i}$, but the clause matches the input $x_{i}$. In such a case, the states of excluded literals are increased in an attempt to change the output of the clause, increasing the discrimination power by introducing literals that invalidate the matching\cite{glimsdal2021coalesced}.

\section{Sparse Tsetlin Machine \\ w/Active Literals} 

A fundamental new input space is required to process sparse data efficiently. The STM operates under the assumption of the sparse data structure, Compressed Sparse Row (CSR). This structure depicts only the active, non-zero values in the data, compressing the input space. However, the STM goes one step further by discarding all negated features. Traditional TM training assumes the expansion of literals to $2o$ features per instance, with literals $x_k$ for $o+1 \leq k \leq 2o$ exclusively representing negated features, effectively doubling the data size at every instance. On the other hand, assuming the data dimensions are known beforehand, lossless compression can be leveraged by discarding negated features and employing CSR. In cases where literals are present in the input, their negated form is implicitly assumed to be absent. Similarly, if a literal is not present in the input, it is understood to be in its negated form. This understanding enables discarding all negated representations, avoiding expansion to $2o$ features, and making the input space more compact, simpler, and efficient. The STM verifies the presence or absence of literals and concludes whether corresponding negated features are in effect. Therefore, input space is more diminutive and compact by leveraging sparsity using an accurate sparse representation by discarding zero values.

\begin{figure}[!h]
    \centering
    \includegraphics[]{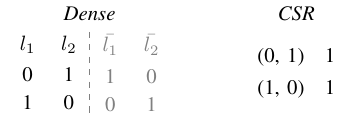}
    \caption{Original dense input and corresponding CSR representation.}
    \label{fig:csr_desc}
\end{figure}

\subsection{Active Literals}

The structure of the STM (Eq.~\ref{eq:stm_structure_tuple}) is similar to that of a CoTM.

\begin{equation}\label{eq:stm_structure_tuple}
    \{ \mathcal{X} , \mathcal{Y} , \mathcal{C} , \mathcal{W} , AL\}
\end{equation}

However, the STM introduces an additional component called Active Literals (AL). The AL functions as a record of literals, effectively working as a gatekeeper for the memory space. Each classification label is assigned its unique AL that records the active literals for a given label. Populating the AL with active literals commences during feedback and is a critical new training phase. Literals are submitted to the AL based on pattern mining.  Consequently, the AL becomes populated with found active literals, effectively establishing an archive of the most significant features observed. If a literal within the AL is deemed sufficient, it will be forwarded to memory.

\begin{equation}
        \begin{aligned}
          AL &= 
            \begin{bmatrix}
              al_{11} & al_{12} & \dots & al_{1a}\\
              al_{21} & al_{22} & \dots & al_{2a}\\
              al_{31} & al_{32} & \dots & al_{3a}\\
               \vdots & \vdots & \ddots & \vdots \\
              al_{m1} & al_{m2} & \dots & al_{ma}\\
              
        \end{bmatrix}\\\\
        &= [al_{ma}]^{m\times a} \in \{l_0, l_1, l_2, \dots, l_o\}
        \end{aligned}
      \label{equation:al1}
\end{equation}

Eq.~\ref{equation:al1} notes how the AL is allocated. AL resources are distributed for each $m$ classification label in $\mathcal{Y}$. This ensures that $al_m \in AL$ uniquely records any label's active literals. The content of AL is an unordered subset of non-negated literals without duplicates, where each active literal $al \in AL$ represents a literal $l_o$ sampled from $\mathcal{X}$. Further, a user-defined parameter, $a$, sets a strict limit on the number of literals stored in the record at any given time, allowing AL to be configured to operate in a static or dynamic mode. Once filled, the AL retains literals in static mode, creating a random-forest-like behavior with static sub-patterns contained in the AL. Conversely, the literals in the AL can fluctuate in dynamic mode. This allows new literals to replace already seen literals within the AL with ones observed from data, thereby providing a steady stream of new literals.

\subsection{Sparse Memory Space}

The sparse clause structure is defined with lists. Novel to the STM, with the introduction of the AL, is the ability for sparse clauses to start empty without any record of existing literals or states. Then, through feedback, literals are introduced to clauses from frequently observed patterns in the AL, populating the clauses with only the essential literals observed from the data. There are two main advantages to this approach. Firstly, instead of initializing a subset \cite{bhattarai2023contracting} or the entire memory all at once \cite{granmo2021tsetlin}, literals are introduced gradually, allocating space only for the features relevant to inference. This not only helps to avoid memory limitations but also ensures that resources are used efficiently. Secondly, this approach makes memory more flexible, as literals can be introduced without requiring pre-defined representations, allowing for the introduction of new features on the go.

\begin{figure}[ht!]
    \centering
    \includegraphics[]{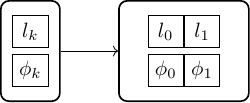}
    \caption{Structure of a sparse clause $C_{j}$.}
    \label{fig:sparse_clause}
\end{figure}

\begin{figure}[!ht]
    \centering
    \includegraphics[]{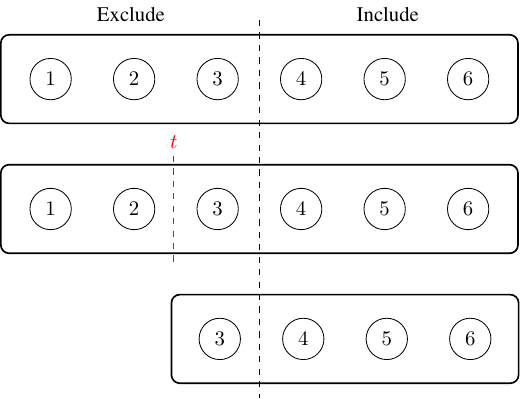}
    \caption{Sparse TAs are defined by assigning a lower threshold $t$ for states. TAs reduced below the allowed threshold get removed from clauses.}   
    \label{fig:removing_below_t}
\end{figure}

In more detail, each clause $C_{j}$ in the sparse representation contains two lists. The first list comprises literals, represented by $l_k$, and the second list represents the states of corresponding TAs, represented by $\phi_k$. The literals list $l_k$ shows which features, out of a total of $o$, are represented by the states $\phi_k$. A sparse state is an integer with $2N$ possible action states. The first $N$ states represent the Exclude action, and the remaining $N+1 \text{ to } 2N$ states represent the Include action. However, a new parameter, $t$, sets a strict lower threshold for a TA state. This parameter creates a configurable spectrum of states, allowing TAs to fluctuate within this spectrum without being removed. If any TA is reduced below the threshold $t$, it is deemed unfit for inference and discarded from the clause. A standard TA and a sparse TA are visualized in Fig. \ref{fig:removing_below_t}. The upper figure depicts a standard TA with $6$ action states. Introducing the lower threshold $t$ limits the states available, resulting in sparse TAs. If any TA is reduced below the threshold $t$, it is deemed unfit for inference and discarded from the clause.

\begin{figure*}[!ht]
    \centering
    \includegraphics[]{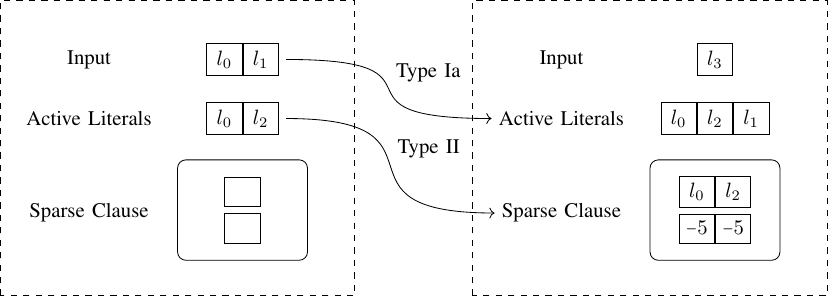}  
    \caption{STMs uses AL as a gatekeeper to introduce literals to clause during feedback.}
    \label{fig:sparse_learning}
\end{figure*}

TAs not present in the clauses, either discarded or not introduced, are not defined, and there is no record of their state or action. With its dynamic introduction and discarding of TAs during training, the new structure enables more efficient training by only distributing feedback to literals deliberately introduced into memory. 

\subsection{Updating the memory space}\label{subsection:updating_clause}

The learning scheme of the STM follows that of the CoTM closely, described in Section~\ref{section:coalesced_description}. In addition, Focused Negative Sampling (FNS) is utilized to better distinguish between classes by forcing similar classes further away from each other \cite{glimsdal2022focused}. Feedback only differs in updating the AL, which occurs exclusively in feedback types Ia and II.

\textbf{Type Ia Feedback} is given to clause $C_j$ if it affirms the input $x_i$ and is assigned to the output value $y_i$, boosting true positives. The STMs AL are populated during Type 1a feedback. Shown in Fig. \ref{fig:sparse_learning}, if a TA were to receive Type Ia feedback, but it is not contained within a clause, that literal would be introduced to the AL for the given class $y_i$. The STMs inherent focus on positive literals means Type 1a feedback is applied with Boosting True Positives (BTP) \cite{granmo2021tsetlin}.

\textbf{Type II Feedback} is given to clause $C_j$ if it matches the input $x$ and is assigned to the output value $\lnot y_i$. This increases the discrimination power of the clause by introducing literals that invalidate the matching, combating false positives. In the STM, new literals are introduced from the AL in Type II Feedback. Fig.~\ref{fig:sparse_learning} describes how literals are introduced to clauses. To increase the discrimination power of a clause, the AL will introduce new literals to invalidate the incorrect matching by submitting new literals from the corresponding AL into memory.

During the training phase, the STM memory space gets populated with literals from the AL. For literals introduced into memory, feedback is distributed traditionally as the CoTM scheme depicts. To prevent a clause from getting overpopulated, a strict limit for the maximum number of literals allowed in clauses is enforced. Defined as $p$, the maximum clause size determines the number of literals introduced to any clause at any given time. Suppose a clause is at its maximum capacity; a literal must be discarded by being reduced below the allowed threshold before allowing new introductions.

\section{Empirical results}

In this chapter, we present the empirical results of the STM. We conducted experiments to confirm the STM's ability to effectively leverage sparse data for accurate predictions. We have organized the results into three sections, each focusing on a different performance aspect. Firstly, we present a series of experiments conducted on various datasets, each with varying degrees of sparsity. We recorded the performance for each case. Secondly, we examine the ability to maintain efficiency across an increasingly sparse problem. We discuss how the STM's performance evolves as the size of the dataset increases and highlight the ability to maintain high performance and efficiency when presented with increasingly sparse problems. Lastly, we employed the STM in a large-scale text corpus application, demonstrating the STM's ability to work in compressed input space and, thus, significantly lower memory usage.

\subsection{Experiments details}

We initialized the STM with the following parameters: Number of clauses ($N$), voting margin ($T$), specificity ($s$), active literal size ($a$), lower-state threshold ($t$), and max clause size ($p$). We measured sparsity as the percentage of non-zero values in any data representation. For the first experiment, we tuned the STM for each dataset. In the second experiment, we increased the vocabulary size of an NLP problem. This was done to expand the data while producing increasingly sparse representations. Finally, we performed five experiments on a large-text corpus. We iteratively increased the total number of documents and vocabulary sizes. Unless otherwise noted, all experiments were conducted on an Intel i5-13600KF @ $3.5GHz$, running single-threaded on Ubuntu 20.04~LTS. 

\subsection{Datasets}\label{subsection:datasets}

The training is conducted on $8$ publicly available datasets.

\begin{itemize}

    \item \textbf{CR}\cite{ding2008holistic} The CR dataset contains customer reviews for sentiment classifications from positive or negative rewires. Here, we apply: $N=6500$, $a=200$, $t=30$, $p=115$, and dynamic AL.
    
    \item \textbf{SUBJ} \cite{pang2004sentimental} is a binary classification dataset containing sentences labeled as either subjective or objective. The STM is applied with: $N=1000$, $a=80$, $t=40$, $p=75$ and dynamic AL.
    
    \item \textbf{MPQA} \cite{wiebe2005annotating} contains articles from different news sources, annotated for opinions. Here, we use: $N=1000$, $a=123$, $t=43$, $p=100$, and dynamic AL.
    
    \item \textbf{SST-2} \cite{socher2013parsing} refers to the binary sentiment classification of the Standford Sentiment Treebank, a corpus with fully labeled parse trees, which allows for complete analysis of compositional effects of sentiment in language. Here, we use: $N=3500$, $a=130$, $t=40$, $p=75$, and dynamic AL.
    
    \item \textbf{MR} \cite{tang2015pte} is a movie review dataset used for binary sentiment analysis. We here employ: $N=1500$, $a=130$, $t=80$, $p=150$, and dynamic AL.
    
    \item \textbf{PC} \cite{ganapathibhotla2008mining} is a binary classification dataset that classifies pros and cons. Here, we apply: $N=3500$, $a=100$, $t=60$, $p=130$, and dynamic AL.
    
    \item \textbf{TUNADROMD}\cite{borah2020malware} The TUNADROMD dataset is a supervised tabular malware detection dataset used to train models on detecting malware attacks. We here employ: $N=3000$, $a=100$, $t=50$, $p=50$, and static AL.
    
    \item \textbf{Amazon Review Data All} \cite{ni2019justifying} is a large-scale text corpus dataset containing product reviews distributed over 5 labels. The specific version of the Amazon Review Data used is the raw composition. This entails all the documents from all the sub-categories without document pruning.  
    
\end{itemize}

When bit-encoded for TM training, these datasets become vastly sparse. Therefore, they indicate how well the STM performs when presented with sparse data.

\bgroup
\def\arraystretch{1.7}
\begin{table}[!h]
\centering
\caption{Performance results for the STM on various datasets. The results reported are the sparsity in $\%$, the best reproducible accuracy in $\%$, and average accuracy in $\%$ from the last $25$ epochs.}
\label{tab:acc-table}
\begin{tabular}{c|c|c|c} 
\hline
\textbf{Dataset} &\textbf{Sparsity}&\textbf{Best}&\textbf{Average}\\\hline
        CR&0.409&83.51&81.49\\
        SUBJ&0.189&90.70&89.40\\
        MPQA&0.112&75.80&74.25\\
        SST-2&0.226&80.72&78.93\\
        MR&0.174&79.08&77.18\\
        PC&0.093&91.13&90.41\\
        TUNADROMD&6.029&99.66&99.31\\\hline
\end{tabular}
\end{table}
\egroup

\subsection{Results}

\begin{figure}[!h]
    \centering
    \includegraphics[]{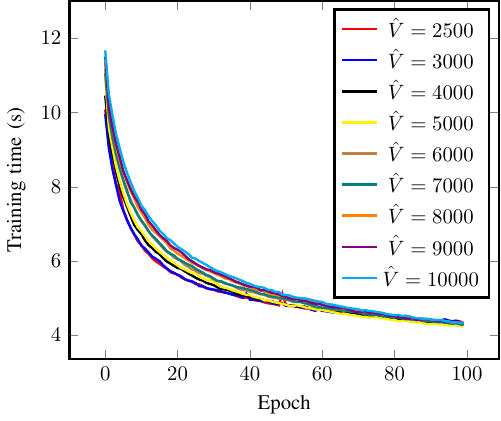}
    \caption{STM training time in seconds (s) per epoch running 100 epochs on the SUBJ dataset with varying vocabulary size $\hat{V}$.}
    \label{fig:plot-time}
\end{figure}
In this section, we present the empirical results of our experimentation. The STM has been rigorously tested on diverse datasets to evaluate its performance. Initially, we focused our evaluation on 6 widely recognized datasets in the field of NLP. These datasets were carefully selected to represent a broad spectrum of NLP tasks and challenges, ensuring a comprehensive assessment of our model’s capabilities. Following the evaluation of NLP datasets, we further extended our experimentation to include a tabular dataset. This allows us to assess versatility and adaptability in handling different data types. Secondly, it provides a more comprehensive understanding of the model’s performance across diverse data domains.

The detailed results of this evaluation are presented in Table~\ref{tab:acc-table}. This table provides a clear and concise summary of the STM's performance on the different datasets tested, attaining competitive results across all the datasets\cite{bhattarai2023contracting}\cite{bhattarai2023tsetlin}\cite{sharma2021human}\cite{sharma2022drop}\cite{borah2020malware}. Noteworthy is that the tabular dataset, TUNADROMD, is the only dataset that benefits from the random-forest-like behavior of the static AL. The TUNADROMD dataset is significantly smaller and less sparse than the NLP datasets. Consequently, it does not benefit from the constant influx of new literals.

\begin{figure}[!h]
    \centering
    \includegraphics[]{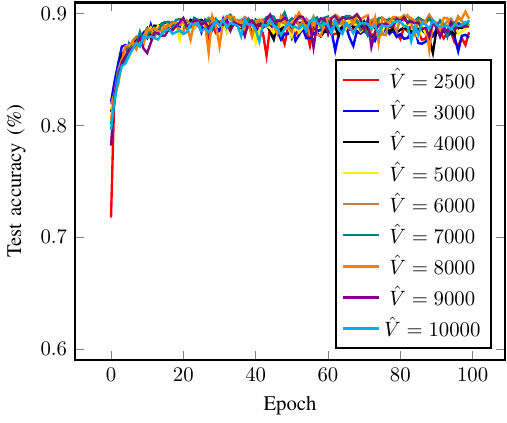}
    \caption{STM test accuracy (\%) per epoch running 100 epochs on the SUBJ dataset with varying vocabulary size $\hat{V}$.}
    \label{fig:plot-acc}
\end{figure}

An experiment was conducted using the SUBJ dataset to verify the STM's sparse data processing capabilities. The experiment involved increasing the vocabulary size, denoted as $\hat{V}$ in $(n \times \hat{V})$, to monitor the performance on increasingly sparse problems. As the vocabulary size increases, the data becomes more complex and high-dimensional, challenging the conventional characteristic of TM training becoming inefficient with large input spaces. The experiment began with a vocabulary size of $\hat{V}=2500$ and was gradually increased by 500 until it reached $\hat{V}=10000$. The results of the experiment are presented in Fig.~\ref{fig:plot-time} and \ref{fig:plot-acc}.

We especially note the training times recorded. From the initial representation with $\hat{V}=2500$ to $\hat{V}=10000$, there is, in effect, a $4\times$ increase in data, distributed in a $49.20\%$ increase in non-zero values and a $301.28\%$ increase in zero values, making the representation vastly more sparse. However, the STM's excellent processing capability of sparse data only results in a minor training time impact. The difference between the two extremes tested resulted in a training time increase of only $15.79\%$ for the first epochs. For the $100th$ epoch, this increase was even less at just $1.47\%$ while achieving higher test accuracy. This confirms the STM's ability with the AL to distinguish the important literals, disregard non-important features, and maintain efficient processing. Initially, literals are introduced into memory. However, as the training progresses, it becomes more adept at recognizing essential features. Consequently, memory is adjusted by discarding non-important literals, highlighting the efficiency of the STM.   

\begin{figure*}[!t]
    \centering
    \includegraphics[]{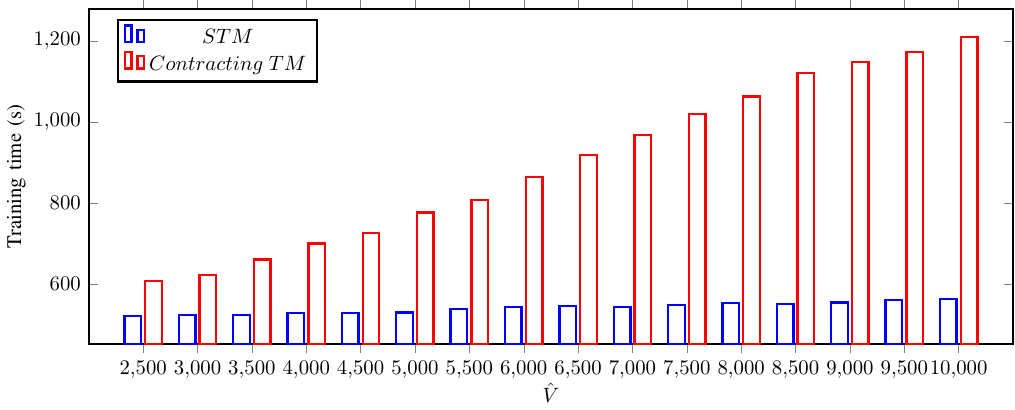}
    \caption{Cumulative training time for STM and Contracting TM over $100$ epochs on the SUBJ dataset with varying vocabulary size $\hat{V}$. The Contrasting TM tuned for $1000$ clauses with an absorbing state of $50$ and literal sampling of $0.6$, while STM applied parameters from section \ref{subsection:datasets}}
    \label{fig:plot-cum}
\end{figure*}

\begin{figure}[!h]
    \centering
    \includegraphics[]{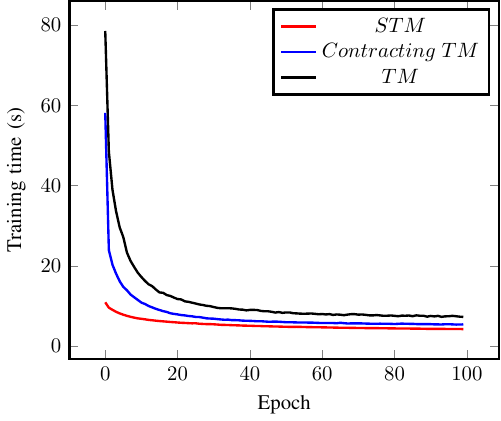}
    \caption{STM, TM, and Contracting TM training time in seconds (s) per epoch running 100 epochs on the SUBJ dataset with $\hat{V} = 5000$.}
    \label{fig:plot-5000}
\end{figure}

Further, we compare the STM to another sparse TM approach, the Contracting TM \cite{bhattarai2023contracting}. We record the cumulative training time from the previous experiments over the $100$ training epochs. Fig. \ref{fig:plot-cum} represents the results from the comparison. As observed in the previous experiments, we note the STM's ability to maintain efficient processing, only resulting in a $8.22\%$ cumulative training time increase from $\hat{V} = 2500$ to $\hat{V} = 10000$. For the Contracting TM, the increase in training time is recorded to be $99.47\%$. We suspect that the Contracting TM's constraint of initializing TM training with literals present in its exclude list makes the early training epochs excessive, and looking closer at the training times per epoch, we can see that this becomes apparent. Fig. \ref{fig:plot-5000} shows the Contracting TM's impressive capability of lowering the training time per epoch during the training phase. However, the requirement of starting with literals present in the exclude list and pruning the memory by absorbing literals scales poorly with increased data sizes and accumulates in an excessive cumulative training time. The STM is not restricted by initializing the memory with literals; instead, it starts with empty memory and employs the AL to introduce sufficient literals, resulting in more efficient training.

\subsection{Deploying STM on large scale text corpus}\label{section:amazon_res}

The STM's excellent processing of sparse data opens up new possibilities for large-scale text corpus applications. Its efficient processing of sparse data and diminutive memory impact make it an ideal solution for handling large volumes of textual data. The Amazon Review All data bank is a large-scale text corpus accumulating $233.1$ million in individual textual documents, representing product reviews with 5 classification labels. We conducted five experiments to explore the STM's scalability by randomly sampling a subset of $n$ documents from the text corpus. Initially, we start with a small subset, gradually increasing the number of documents $n$ and vocabulary sizes $\hat{V}$ to stress the STM. The experiment was conducted using traditional TM training without batching.

Table \ref{tab:amazon-table} depicts the outcomes of the experiments. The STM displays that it can effectively scale with the data size, improving performance as document and feature counts increase. Moreover, we also analyze the memory requirements associated with the experiments, observing that as the data dimensions $(n \times \hat{V})$ grow, the corresponding memory impact escalates. Later experiments necessitate terabytes of system memory to traditionally allocate the data adequately. In addition, the STM's unique ability to set strict limits on memory space, with $a$, $p$, and $t$, ensures that the TM memory impact remains consistent. Combined with improved processing of sparse data and compressed input space, this enables TM training in applications previously considered practically unfeasible.

\section{Conclusion}
In this paper, we introduced the Sparse Tsetlin Machine (STM), a novel approach for efficiently processing sparse data. By leveraging Active Literals, the STM represents a true sparse representation of the Tsetlin Machine that reduces memory impact and increases sparse data processing capabilities. The STM has been rigorously evaluated and is supported by competitive results on 8 well-known datasets. It can determine what literals are sufficient for training, therefore not adhering to traditional TM characteristics of inefficient scalability. Additionally, we experimented with the STM in large-scale text corpus applications and showed how the STM can be applied in training regiments previously unavailable.

\bgroup
\def\arraystretch{1.65}
\begin{table*}[!t]
\centering
\caption{Performance results for the STM on the Amazon Review Data trained for $10$ epochs on 20 CPU threads. Results reported are the number of documents assigned training and testing, vocabulary sizes, system memory required to allocate the bag-of-word processed data represented as $uint8$ without CSR, dense TM and sparse STM memory impact, the mean epoch time in seconds, and the best reproducible test accuracy in $\%$}
\label{tab:amazon-table}
\begin{tabular}{c|c|c|c|c|c|c|c}
\hline
Train Documents & Test Documents & Vocabulary Size & Data size & Dense Memory & Sparse Memory & Epoch Time 
& Test Accuracy \\ \hline
        80,000 & 20,000 & 250,000& 23.3 GiB& 953.67 MiB & 0.544 MiB &102.6 &60.16\\
        800,000&200,000 &1,000,000&931 GiB& 3814.6 MiB &0.544 MiB &1007.4 &61.45\\
        2,000,000&500,000&2,500,000&4.55 TiB& 9536.74 MiB &0.544 MiB &2392.9 &64.19\\
        4,000,000&1,000,000&2,500,000&11.4 TiB& 9536.74 MiB &0.544 MiB &4133.7 &66.27\\
        8,000,000&2,000,000&5,000,000&45.5 TiB& 18.62 GiB &0.544 MiB &9916.2 &67.31\\ \hline

\end{tabular}
\end{table*}
\egroup

\section*{Acknowledgment}
This paper is partly funded by Forsta. We are grateful for their support.

\bibliographystyle{IEEEtran}
\bibliography{bibliography}

\end{document}